\documentclass[letterpaper]{article} 
\usepackage{aaai25}    
\usepackage{times}  
\usepackage{helvet}  
\usepackage{courier}  
\usepackage[hyphens]{url}  
\usepackage{graphicx} 
\usepackage{natbib}  
\usepackage{caption} 
\usepackage{booktabs} 
\usepackage{array}
\usepackage{xcolor}

\usepackage{algorithm}
\usepackage{algorithmic}

\usepackage{newfloat}
\usepackage{listings}
\DeclareCaptionStyle{ruled}{labelfont=normalfont,labelsep=colon,strut=off} 
\floatstyle{ruled}
\newfloat{listing}{tb}{lst}{}
\floatname{listing}{Listing}
\title{SEGMN: A Structure-Enhanced Graph Matching Network for Graph Similarity Learning}

\author{ 
    Wenjun Wang\textsuperscript{\rm 1}, Jiacheng Lu\textsuperscript{\rm 1}, Kejia Chen\textsuperscript{\rm 1,*}, Zheng Liu\textsuperscript{\rm 1}, Shilong Sang\textsuperscript{\rm 1} 
}
\affiliations{ 
    \textsuperscript{\rm 1}Nanjing University of Posts and Telecommunications, Nanjing, China\\
    \{b21030816, b21060620, chenkj, zliu, 1022040812\}@njupt.edu.cn
}

\begin{document}

\maketitle

\begin{abstract}
Graph similarity computation (GSC) aims to quantify the similarity score between two graphs. Although recent GSC methods based on graph neural networks (GNNs) take advantage of intra-graph structures in message passing, few of them fully utilize the structures presented by edges to boost the representation of their connected nodes. Moreover, previous cross-graph node embedding matching lacks the perception of the overall structure of the graph pair, due to the fact that the node representations from GNNs are confined to the intra-graph structure, causing the unreasonable similarity score. Intuitively, the cross-graph structure represented in the assignment graph is helpful to rectify the inappropriate matching. Therefore, we propose a structure-enhanced graph matching network (SEGMN). Equipped with a dual embedding learning module and a structure perception matching module, SEGMN achieves structure enhancement in both embedding learning and cross-graph matching. The dual embedding learning module incorporates adjacent edge representation into each node to achieve a structure-enhanced representation. The structure perception matching module achieves cross-graph structure enhancement through assignment graph convolution. The similarity score of each cross-graph node pair can be rectified by aggregating messages from structurally relevant node pairs. Experimental results on benchmark datasets demonstrate that SEGMN outperforms the state-of-the-art GSC methods in the GED regression task, and the structure perception matching module is plug-and-play, which can further improve the performance of the baselines by up to 25\%.
\end{abstract}

%

\section{Introduction}
Graph similarity learning refers to learning the similarity score between two graphs, having extensive applications, such as code detection \cite {dai2023study,miss31}, molecular graph similarity \cite{miss1} and image matching \cite{miss2}. Graph Edit Distance (GED) \cite{miss4} and Maximum Common Subgraph (MCS) \cite{miss5} are two widely used measurements,  where GED refers to the minimum number of operations that convert one graph to the other and MCS is the largest subgraph which is simultaneously isomorphic to both graphs. However, the computation of GED and MCS is attributed to NP-complete problem \cite{miss6,miss7}. Traditional algorithms like Hungarian \cite{miss8} and A* \cite{miss9} can compute GED accurately but at the cost of high computational complexity.  

The prevailing of Graph Neural Networks (GNNs) \cite{miss10} promotes the development of deep graph similarity learning \cite{miss29}. Early studies generally learn graph-level representation for comparison \cite{miss17}. To achieve higher granularity matching through node-level comparison, subsequent methods adopt GNNs for node embedding, followed by various cross-graph matching strategies, which can be generally divided into two categories. One is cross-graph attention \cite{miss13,miss14,miss32} that captures abundant node-graph interactions to provide each node in one graph with the other graph's node representations. The other is to directly extract features of the similarity matrix by comparing the similarity score of each cross-graph node pair \cite{miss11,miss19,miss15}. Moreover, there appear a few methods constructing structure-enhanced representations for structure matching. Some adopt the attention mechanism including distance information to enhance node embeddings \cite{miss15}, and others encode edges to directly represent the connection condition \cite{miss16,miss20}. However, the edge embedding is only for graph-level representation or subgraph construction.

  	Despite various matching strategies, existing methods have two main drawbacks. (1) \textbf{Representation limitation.} A comprehensive representation enables multi-view matching. Most methods only use simple node embeddings without highlighting the edge representation, which is crucial for structure matching. (2) \textbf{Matching inadequacy.} Most methods conduct diverse cross-graph matching strategies on node embeddings from GNN. However, the node embeddings are confined to intra-structures as message passing is applied to their respective graphs. Therefore, directly comparing cross-graph node embeddings lacks the matching perception of the whole graph pair structure,  thus causing the unreasonable similarity score. Intuitively, the cross-graph structure relationship provides the inter-graph structure relationship and helps rectify the unreasonable similarity score. According to the cross-graph structure relationship, two cross-graph node pairs are believed structurally relevant when their corresponding nodes in the same graph are adjacent. The concrete cross-graph structure relationship can be described by the assignment graph detailed in Definition 3.   
  	
  	A toy example is given in Fig. 1. In this example, the nodes of $G_1$ and $G_2$ are embedded in a latent space by a GNN, where node $v^1$ is close to node $v^a$ in the space as they are structurally similar. Their respective neighbor nodes $v^3$ and $v^c$ should be relatively close from the view of the whole structure of the graph pair while they are initially not. Cross-graph node pair $(v^3, v^c)$ and $(v^1, v^a)$ are structurally relevant as $v^1$ and $v^3$, $v^a$ and $v^c$ are both adjacent in $G_1$ and $G_2$. Based on the cross-graph structure relationship, the similarity score of  $(v^3, v^c)$ can be increased by simply aggregating the similarity score of  $(v^1, v^a)$, which is quite high. So the unreasonable similarity score of $(v^3, v^c)$ can be rectified.

 \begin{figure}
  \centering
  \includegraphics[scale=0.125]{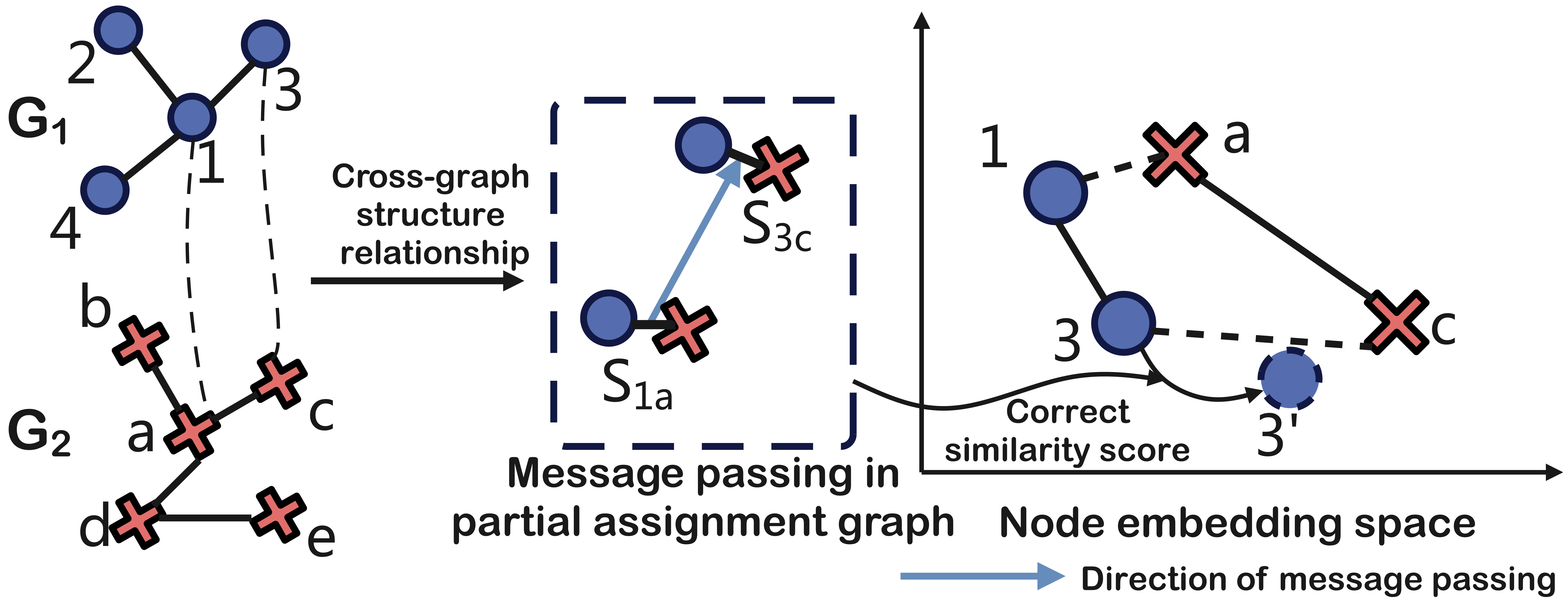}
  \caption{An example of cross-graph structural correction. The figure shows a pair of graphs with their partial assignment graph and node embedding space. The closer the two embeddings are in the space, the more similar they are supposed to be. $S_{1a}$, $S_{3c}$ represent the similarity scores of node pair $(v^1, v^a)$ and $(v^3, v^c)$.}
  \label{fig:small-image}
\end{figure}

To overcome the above drawbacks in previous works, we design a structure-enhanced framework, which achieves structure matching from the perspective of both representation and matching. On one hand, we propose a dual embedding approach that adds the structure perspective from the adjacent edges of each node. Firstly, the node graph is transformed into its corresponding line graph (detailed in Definition 2) with edge features. Then, an edge-enhanced GCN is applied to the line graph to learn node embedding (\textit{i.e.}, edge embedding in the node graph). Finally, each node embedding is concatenated to the aggregated embeddings of its adjacent edges in the node graph to generate the ultimate dual embedding. On the other hand, we introduce a structure perception matching module based on the structure-enhanced algorithm to augment the cross-graph similarity matrix. The assignment graph is adopted to obtain the neighbor relationships of each cross-graph node pair in their respective graph, and the similarity score of each node pair is convolved with that of its adjacent node pairs in the assignment graph to enhance the structure matching between graphs.

  	  Our contributions can be summarized as follows:

  \begin{itemize}
    \item A structure-enhanced graph matching framework is proposed, which fully utilizes the graph structure from the perspective of both representation and matching.
    \item Instead of aggregating messages only from nodes, a dual embedding learning module is proposed which can additionally aggregate adjacent edge representation into each node for a structure-enhanced representation.
  
    \item Different from previous methods, our proposed structure perception matching module introduces the cross-graph structure to enhance matching by aggregating the similarity score on the assignment graph. It is a plug-to-play module that can be applied to most GSC methods.
    
   \item Experiments on the benchmark datasets show that our method outperforms the state-of-the-art methods. The effectiveness of the proposed modules is also verified.
  \end{itemize}
\section{Related Work}

\noindent \textbf{Cross-graph interaction.} With the success of GNNs, graph similarity learning has been well explored from the graph-level embeddings \cite{miss17} to a more subtle node-level comparison. SimGNN \cite{miss11} firstly uses a similarity matrix to extract node-level features through histogram. However, the effect is limited as histogram is non-differentiable. For better cross-graph interaction, subsequent methods are divided into two categories. One is extracting features from the cross-graph similarity matrix, such as GraphSim \cite{miss19}, DeepSIM \cite{miss30}, and NA-GSL \cite{miss15}. They calculate the similarity score of cross-graph node pair to construct similarity matrices and extract their features by aggregation or specially designed CNNs. The other is to attain cross-graph node embedding through cross-graph attention. GMN \cite{miss12} proposes a Graph Matching Network that aggregates messages from both graphs through cross-graph attention for the initial node embeddings. While MGMN \cite{miss13}, CGMN \cite{miss14}, and GED-CDA \cite{miss28} all adopt cross-graph attention after generating node embeddings. By calculating the similarity scores of cross-graph node pairs as attention coefficients, these methods can incorporate the node embeddings from the other graph.  
   
 Despite the success of previous methods, the node embeddings for cross-graph interaction are confined to intra-graph structures and lack the view of the inter-graph structures.
 
 \noindent \textbf{Structure-enhanced representation.} To better extract graph structures for matching, edge embedding is suited as the connection condition naturally describes the graph’s structure. Recently, there appear a few graph similarity learning methods that incorporate edge-level representation. DGE-GSIM \cite{miss16} uses edge embedding only to construct graph-level matching. ISONET \cite{miss20} proposes an edge alignment network to learn the optimal alignment of edges in two graphs. But the edge embeddings are not used for graph structure matching, but to build subgraphs. To enhance embedding with structure information, there appear some other methods. MATA* \cite{miss33} enhances node embedding with additional structure information by conducting random walk in graphs. Compared to GNN, the representation ability of random walk is relatively weak. NA-GSL \cite{miss15} designs a self-attention module that compares the distance between node pairs to indirectly consider the structure relationship.
 
 \begin{figure*}
  \centering
  \includegraphics[width=0.95\textwidth]{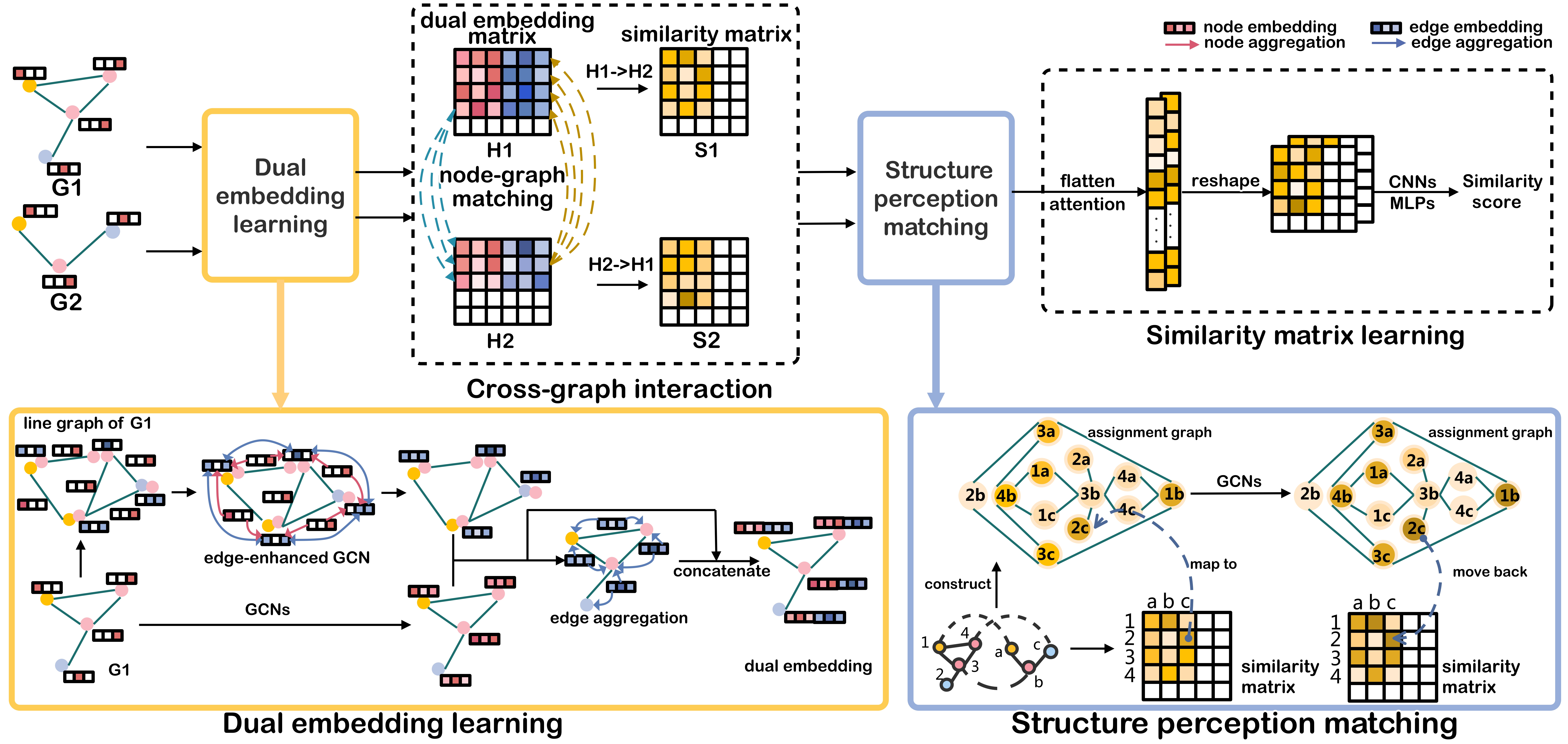}
  \caption{Framework of SEGMN.}
  \label{fig:example}
\end{figure*}

To generate better edge embeddings, we propose an edge-enhanced GCN on the line graph of its original graph, and each node embedding will be concatenated to the aggregated edge embeddings for more efficient edge-level matching.

\section{Preliminaries}
In this section, node graph, line graph, and assignment graph are defined. Node graph is the ordinary graph. Line graph is the graph transformed from the node graph. Assignment graph \cite{miss23, miss36} shows the connection condition of each cross-graph node pair, which describes the cross-graph structure relationship.

    \noindent \textbf{Definition 1 : Node graph.} An undirected node graph \textit{G = (V, E, A, X, Y, K)} is composed of the node set 
\textit{V} with $\left| V \right| = N$, edge set \textit{E} with \textit{$\left|E \right| = M$}, adjacency matrix $A \in R^{N \times N}$, node feature matrix $X \in R^{N \times d_1}$, edge feature matrix $Y \in R^{M \times d_2}$, and incidence matrix $K \in R^{N \times M}$. Here, $K$ describes the connection relationship between edges and nodes. If edge $e^k$ connects with node $v^i$ and node $v^j$, then $K_{i, k}$ and $K_{j, k}$ is 1, else 0.

	\noindent \textbf{Definition 2 : Line graph.}  
	Given the node graph $G$, $G_{E}=(V_{E}, E_{E}, A_{E}, X_{E}, Y_{E})$ is its line graph. A given node $v^{ij} \in G_{E}$ corresponds to the edge connecting the node pair $(v^i, v^j)$ in $G$, and the feature of $v^{ij}$ in $G_{E}$ is attained by adding or concatenating the node features $v^i$ and $v^j$ in $G$. If edge $e^i$ is adjacent to edge $e^j$ in $G$, their corresponding nodes are connected in $G_E$. If two edges are connected through node $v^i$ in $G$, the feature of the edge between their corresponding nodes in $G_ {E}$ is exactly the feature of $v^i$ in $G$. A concrete example is in Appendix B.3.

	\noindent \textbf{Definition 3 : Assignment graph.} 
	Given node graphs $G_1= (V_1,E_1)$, $G_2= (V_2,E_2)$, the assignment graph $G_{A}= (V_{A},E_{A})$ is constructed as foll ows. For node construction, cross-graph node pair $(v^i_1,v^a_2) \in V_1 \times V_2$ is taken as a node $v^{ia}_A \in V_{A}$, and each cross-graph node pair is constructed to be a node in $G_A$, similarly. For edge construction, there exists an edge between node $v^{ia}_{A}$ and node $v^{jb}_{A}$ in $G_{A}$ if and only if both edges exist in its original graph, $i.e.,(v^i_1,v^j_1) \in E_1$ and $(v^a_2,v^b_2) \in E_2$. A concrete example is in Appendix B.4.

\section{Method}

\subsection{Overview}
The proposed structure-enhanced graph matching network (SEGMN)  consists of four modules, as shown in Fig. 2.
\noindent  	\textbf{(1) Dual embedding learning}. An edge-enhanced GCN is applied on the line graph $G_{E} $ for edge embeddings in node graph $G$, and then each node embedding in $G$ is concatenated to its adjacent edge embeddings for dual embeddings. 
     \noindent 	\textbf{(2) Cross-graph interaction}. The similarity score between cross-graph dual embedding pairs is calculated with a node-graph attention mechanism. 
     \noindent 	\textbf{(3) Structure perception matching}. The convolution on assignment graph $G_{A}$ helps enhance the similarity matrix for structural matching. 
     \noindent  	\textbf{(4) Similarity matrix learning}. The self-attention and CNNs are applied to extract features of the similarity matrix.

\subsection{Dual embedding learning}
The dual embedding learning module can be divided into three steps. Firstly, an edge-enhanced GCN is applied to the line graph for edge embedding. Secondly, a GCN with residual connections is applied to the node graph for node embedding. Thirdly, each node aggregates the connected edge embeddings to generate the ultimate dual embedding.

\noindent \textbf{Edge embedding learning.} Firstly, the node graph $G$ is transformed to its corresponding line graph $G_{E}$, then an edge-enhanced GCN is conducted to get node embeddings $H^{v}_{{E}}$ in $G_{E}$.  Specifically, each node receives messages from both adjacent nodes and connected edges:
\begin{equation}
    H^{v}_{E}= H^{v \leftarrow v}_E+ H^{v \leftarrow e}_{E}
\end{equation}
where $H^{v}_{E}$ denotes the overall node embedding in $G_{E}$, $H^{v \leftarrow v}_{E}$ denotes the embedding aggregated from neighbor nodes and $H^{v \leftarrow e}_{E}$ denotes the embedding aggregated from adjacent edges.

For node message propagation,  the GCN with residual connection is  adopted to generate the node embedding:
\begin{equation}
    {H^{v \leftarrow v}_{E}}^{(l+1)}=\sigma(\widetilde{D}^{-\frac{1}{2}}_{E} {\widetilde{A}}_{E} \widetilde{D}^{-\frac{1}{2}}_{E} {H_{E}^{v \leftarrow v}}^{(l)} W^{(l)}_{E})+{H^{v \leftarrow v }_{{E}}}^{(l)}
\end{equation}
where $\widetilde{A}_{E}=A_{E}+I_{E}$ is the adjacency matrix adding self-connections, $\widetilde{D}_{E}$ is the degree matrix, and  $W^{(l)}_{E}$ is the parameter matrix.

For edge message propagation,  the parameterized incidence matrix multiplied with transformed edge features is used to boost the node representation, summarized as:
\begin{equation}
    {H^{v \leftarrow e}_{{E}}}^{(l+1)}=(K_{E} \odot W_1 )\cdot tanh(Y_{E} W_2 + b)
\end{equation}
where $K_{E}$ denotes the incidence matrix of $G_{E}$, $W_1$ refers to the learnable parameter matrix which decides the influence from all connected edges by calculating the Hadamard product with $K_{E}$ and applying the linear transformation to edge features $Y_{E}$, $W_2$ refers to the parameter matrix and $b$ is the bias. Considering that the influence from edges can be positive or negative, we apply the $tanh$ function to describe the uncertainty \cite{miss21}. Thus, each node representation can be improved by positive or negative influence with diverse weights of its connected edges.

As each node in $G_{E}$ corresponds to one edge in $G$, the node embedding $H^{v}_{E}$ in $G_E$ can be converted  to be the edge embedding in $G$,  notated as $H^e$:
\begin{equation}
    H^{e}= H^{v}_{E}
\end{equation}

\noindent \textbf{Node embedding learning.} The node embedding learning is based on node graph $G$. We apply the GCN consistent with Eq. (2) for node embeddings $H^{v}$ in $G$:
\begin{equation}
    H^{v^{(l+1)}}=\sigma(\widetilde{D}^{-\frac{1}{2}} \widetilde{A}\widetilde{D}^{-\frac{1}{2}} H^{v^{(l)}} W^{(l)})+H^{v^{(l)}}
\end{equation}

where the notations are similar to that in Eq. (2) but belong to graph $G$. After several layers of convolution, the node embedding is obtained and denoted as $H^{v}$.

\noindent \textbf{Dual embedding generating.}  It is time-consuming to conduct edge-level cross-graph matching for graphs with considerable edges. So we take the strategy of dual embedding. Specifically, the dual embedding $H^{dual}$ in node graph $G$ is the concatenation of the node embedding $H^{v}$ and the aggregation of its adjacent edge embeddings (referring to $H^{e}$) so that the graph similarity can be evaluated from the view of both the node and its connected edges to achieve structure-enhanced matching.
\begin{equation}
    H^{dual}= H^{v} \oplus ( {K'} \cdot H^{e})
\end{equation}
where $\oplus$ is the concatenation operation, and $K'$ denotes a modified incidence matrix of $G$. In the initial incidence matrix $K$, $K_{i, k}$ and $K_{j, k}$ are 1 if edge $e^k$ connects with node $v^i$ and node $v^j$. When aggregating the message of edges, each edge has the same weight, which is unreasonable. Instead,  $K_{i, k}$ and $K_{j, k}$ are modified to get $K'_{i, k}$ and $K'_{j, k}$ in $K'$:
\begin{equation}
	K'_{i, k}=K'_{j, k}={(d_i)}^{-\frac{1}{2}} {(d_j)}^{-\frac{1}{2}}
\end{equation}
where $d_i$ and $d_j$ represent the degree of $v_i$ and $v_j$, respectively. The new incidence matrix $K'$ implies not only the edge connection condition but also the structure information presented by degree, so the structural influence from different edges is aggregated to each node by weighting.

\subsection{Cross-graph interaction}
The cross-graph interaction is conducted for node-level matching. Here,  the node-graph attention is adopted to calculate each node pair's similarity score.

Firstly, a linear transformation is used to generate a query and key for each graph's dual embedding:
\begin{equation}
    Q_i=H^{dual}_i W^q_i, K_i=H^{dual}_i W^k_i \\
\end{equation}
where $i \in \{1,2\}$.

Then, each node embedding in one graph's query matrix is compared with all of the node embeddings in the other graph's key matrix. The scaled dot product is used as the similarity function. The attention coefficient is conducted twice for two different perspectives of similarity between two graphs, so two similarity matrices are obtained:

\begin{equation}
    S_1= \mathrm{softmax}(\frac{Q_1 (K_2)^T}{\sqrt{d_k}}),S_2= \mathrm{softmax}(\frac{Q_2 (K_1)^T}{\sqrt{d_k}})
\end{equation}

Finally, each similarity matrix is filled with zero into the dimension of $N_{max} \times N_{max}$, where $N_{max}$ refers to the maximum node number of the graphs. It makes the similarity matrix flexibly adapt to different graph sizes, and the mask mechanism is applied to ignore the filled part.

\subsection{Structure perception matching}
The structure perception matching is proposed based on the structure-enhanced algorithm, which applies cross-graph structure relationships to help enhance the similarity matrix.

\noindent \textbf{Structure-enhanced algorithm.} The main idea is that each cross-graph node pair finds its structurally related cross-graph node pairs through function $GenStructure$ ( · , · ), and then aggregates their messages through function $Aggregate$ ( · ). Here the message of a node pair is exactly its similarity score.  

\begin{algorithm}[tb]
\caption{Structure-enhanced Algorithm}
\label{alg:structure_enhanced_algorithm}
\textbf{Input}: $G_1 = (V_1, E_1)$, $G_2=(V_2,E_2)$, with $\left| V_1 \right| = N_1$, $\left| V_2 \right| = N_2$, the similarity matrix $S$ between $G_1$ and $G_2$\\
\textbf{Output}: The structure-enhanced similarity matrix $S'$
\begin{algorithmic}[1] 
\FOR{$i \in [1, N_1]$}
    \FOR{$j \in [1, N_2]$}
        \STATE $set^{ij} \leftarrow$ GenStructure$(G_1, G_2)$
        \STATE $message \leftarrow$ Aggregate($S_{set^{ij}}$)
        \STATE $S'_{ij} \leftarrow message$
    \ENDFOR
\ENDFOR
\STATE \textbf{return} $S'$
\end{algorithmic}
\end{algorithm}

 \begin{figure}
  \centering
  \hspace{-0.7cm}
  \includegraphics[scale=0.2]{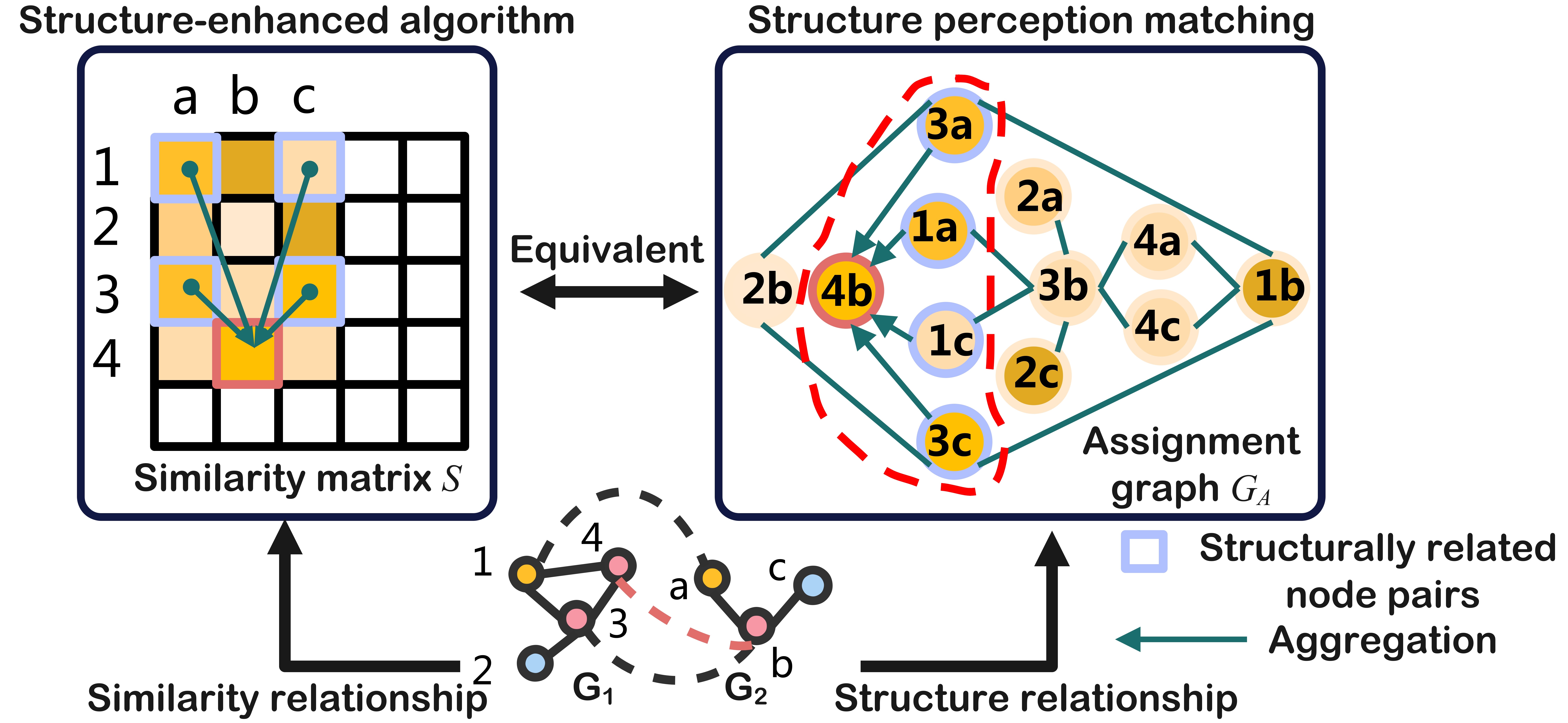}
  \caption{Relationship between structure perception matching and structure-enhanced algorithm. }
  \label{fig:small-image}
\end{figure}

\floatplacement{algorithm}{top}

The detailed algorithm is in Algorithm 1, where $set^{ij}$ refers to the set of node ID pairs whose node pairs are structurally related with node pair $(v^{i}_1,v^j_2 )$. For example, if node pair $(v^{m}_1,v^n_2)$ and $(v^{p}_1,v^q_2)$ are structurally related to  $(v^{i}_1,v^j_2 )$, then $set^{ij}= \{ (m,n), (p,q)\}$. $S_{set^{ij}}$ refers to taking out the similarity scores from $S$ based on $set^{ij}$, which is exactly the similarity scores of node pairs in $set^{ij}$ (\textit{i.e.} $S_{mn}$, $S_{pq}$ here).

For function $GenStructure$ ( · , · ),  the operation of finding node pairs that are structurally related with node pair $(v^{i},v^j )$ is exactly finding the neighbor nodes of $v^{ij}_A$ in assignment graph $G_A$. For function $Aggregate$ ( · ), it can be common aggregating functions like averaging, maximum, MLP, attention mechanism, and so on.

\noindent \textbf{More Details.} The structure perception matching  takes assignment graph $G_A$ to construct function $GenStructure$ ( · , · ) and uses message passing in  $G_A$ to construct $Aggregate$ ( · ). As shown in  Fig. 3, the operation of aggregating neighbor messages in $G_A$  equals to the structure-enhanced algorithm. Take the node $v^{4b}$ in $G_A$ as an example, the neighbors of $v^{4b}$ are the output of function $GenStructure$ ( · , · )  in the structure-enhanced algorithm, and passing their messages to $v^{4b}$ equals to aggregating message in the similarity matrix $S$. Therefore, the similarity score can be mapped into the node feature of $G_{A}$ to be processed with GNN.

Firstly, the feature of $v^{ia}$ in $G_{A}$ is  the corresponding node pair's similarity score in the similarity matrix:
\begin{equation}
    x_{{A}}^{ia}= s^{ia}
\end{equation}
where $x_{{A}}^{ia}$ refers to the feature of node $v^{ia}$ in $G_{A}$, and $s^{ia}$ refers to the similarity score of a cross-graph node pair $(v^i_1,v^a_2) \in V_1 \times V_2$. 

Then, a vanilla GCN is used for $G_{A}$ so that each similarity score is convolved with the similarity scores of its neighbors:
\begin{equation}
    \widetilde{X}_{A}=\sigma(\widetilde{D}^{-\frac{1}{2}}_{A}\widetilde{A}_{A}\widetilde{D}^{-\frac{1}{2}}_{A} X_{A} W_{A})
\end{equation}

Finally, the node features in the assignment graph $G_A$ are moved back to the similarity matrices.
\begin{equation}
    (s')^{ia}=\widetilde{x}_{{A}}^{ia}
\end{equation}
where $\widetilde{x}_{{A}}^{ia}$ denotes the updated feature of node $v^{ia}$ in $\widetilde{X}_{A}$. The structure-enhanced matrices are denoted as $S'_1$ and $S'_2$. 

\begin{table*}[!ht]
    \centering
    \small 
    \renewcommand{\arraystretch}{0.95} 
    \setlength{\tabcolsep}{0.3pt}
    
    \label{tab:booktable}
    \begin{tabular*}{1.05\linewidth}{@{\extracolsep{\fill}}lcccc|cccc|cccc|cccc}
        \toprule
        Dataset & \multicolumn{4}{c}{AIDS700} & \multicolumn{4}{c}{LINUX} & \multicolumn{4}{c}{IMDB-small} & \multicolumn{4}{c}{IMDB-large} \\
        
        Metrics  & MSE $\downarrow$ & $\rho$ $\uparrow$ & $\tau$ $\uparrow$ & p@10 $\uparrow$  & MSE $\downarrow$ & $\rho$ $\uparrow$ & $\tau$ $\uparrow$ & p@10 $\uparrow$ & MSE $\downarrow$ & $\rho$ $\uparrow$ & $\tau$ $\uparrow$ & p@10 $\uparrow$  & MSE $\downarrow$ & $\rho$ $\uparrow$ & $\tau$ $\uparrow$ & p@10 $\uparrow$\\
        \midrule
        GCN &5.551 &0.814 &0.657 & 0.151 &3.401 &0.964 &0.896 &0.909 &2.729 & 0.938 &0.877 &0.896  &16.079 &0.634 &0.461 &0.210\\
        GIN &4.836 &0.838 &0.687 &0.140  &3.057 & 0.963 &0.895 &0.782 &1.875 &0.928 &\underline{0.909} &0.907 &15.879 &0.625 &0.443 &0.350\\
        GAT &5.339 &0.828 &0.673 &0.151 &4.148 &0.945 &0.858 &0.893 &1.206 &0.953 &0.896 &0.956 &15.938 &0.624 &0.453 &0.310\\
        \midrule
        SimGNN & 1.573 & 0.835 & 0.678 & 0.417  & 2.479 & 0.912 & 0.791 & 0.635 & 1.104 & 0.887 & 0.779 & 0.958 & 6.494 & 0.703 & 0.554 & 0.556\\
      
        GraphSim & 2.014 &0.839 &0.662  &0.401 & 0.762 &0.953 &0.882 &0.956 &0.723 &0.940 &0.881 &0.976 &4.441 &0.763 &0.609 &0.603\\
                  
        EGSC-T & 1.601 & 0.901 & 0.739 & \underline{0.658} & 0.163 & 0.988 & 0.908 & \textbf{0.994} & 0.518 & 0.932 & 0.828 & 0.977 & 4.123 & 0.827 & 0.669 & 0.636\\
                  
        MGMN & 2.297 & 0.904 & 0.736 & 0.456 & 2.040 & 0.965 & 0.858 & 0.956 & 0.696 & \underline{0.958} & \textbf{0.915} & 0.971  & 5.472 & 0.760 & 0.614 & 0.538\\
     
        ERIC & \underline{1.383} & 0.906 & 0.740 & \textbf{0.679} & 0.113 & 0.988 & 0.908 & \textbf{0.994} & 0.452 & 0.927 & 0.827 & 0.977  & 2.807 & 0.806 & 0.652 & 0.610 \\
                  
        NA-GSL & 1.852 & 0.890 & 0.745 & 0.510 & 0.133 & \textbf{0.994} & \textbf{0.968} & \underline{0.992} & 0.720 & 0.937 & 0.880 & 0.973 & 4.139 & 0.826 & 0.661 & 0.630\\
        \midrule
        SEGMN(GCN) & 1.386 & \underline{0.910} & \underline{0.773} & 0.597 & \textbf{0.094} & \textbf{0.994} & \textbf{0.968} & \textbf{0.994} & 0.399 & 0.956 & 0.896 & \underline{0.981} & 2.139 & \underline{0.855} & \underline{0.716} & \underline{0.745} \\
        
        SEGMN(GIN) & 1.417 & 0.908 & 0.771 & 0.590 & 0.116 & \underline{0.993} & \underline{0.960} & 0.984 & \underline{0.394} & \underline{0.958} & 0.899 & 0.978 & \textbf{1.999} & \underline{0.855} & 0.713 & 0.723 \\
        
        SEGMN(SAGE) & \textbf{1.370} & \textbf{0.914} & \textbf{0.780} & 0.634 & \underline{0.101} & \underline{0.993} & \underline{0.960} & 0.988 & \textbf{0.360} & \textbf{0.959} & 0.903 & \textbf{0.989} & \underline{2.122} & \textbf{0.877} & \textbf{0.742} & \textbf{0.750} \\
        \bottomrule
        
    \end{tabular*}
    \caption{Evaluation on benchmarks. SEGMN and the state-of-the-art methods are evaluated and compared in terms of MSE ($10^{-3}$), $\rho$, $\tau$ and p@10. The best result is bold and the second-best result is underlined.}
\end{table*}

\begin{table}
    \centering
    \small
    \renewcommand{\arraystretch}{0.95}
    \setlength{\tabcolsep}{1mm}
        \label{tab:booktable}
    \begin{tabular}{lcccccc}
        \toprule
        & \#graphs & $\overline{\left| V \right|}$ & $\overline{\left| E \right|}$ & ${\left| V \right|}_{max}$ & ${\left| E \right|}_{max}$  & labeled\\
        \midrule
        AIDS700       & 700  & 8.9 & 8.8  & 10 & 14  & Yes\\
        LINUX         & 1000 & 7.6 & 6.9  & 10 & 13  & No\\
        IMDB          & 1500 & 13  & 65.9 & 89 & 1467 & No\\
        IMDB-small    & 900  & 8.5 & 27.9 & 12 & 55  & No\\
        IMDB-large    & 300  & 19  & 89.5 & 50 & 435  & No\\
        \bottomrule
    \end{tabular}
    \caption{Statistics of graph datasets.}

\end{table}

\subsection{Similarity matrix learning}
The final similarity score can be predicted by extracting features of the enhanced similarity matrix. 

Firstly, a self-attention mechanism is applied to both similarity matrices. For each similarity matrix, it is flattened into the vector $S'_{flatten}$ with the length of $N_{max} \times N_{max}$ for the self-attention as follows.

\begin{equation}
    S_{flatten}= \mathrm{softmax}(\frac{Q_{flatten} (K_{flatten})^T}{\sqrt{d_k}}) V_{flatten}
\end{equation}
where $Q_{flatten},K_{flatten},V_{flatten}$ are all the linear transformation of  $S'_{flatten}$. 
The updated  $S_{faltten}$ is then mapped back to be the similarity matrix $S'$. 

Then, considering the unique character of the similarity matrix, that is, one column and one row both represent the node-graph similarity vector,  the CNN with cross-shaped filters \cite{miss15} is applied to extract the node-graph similarity vector of node pair $v^{ij}$. For each similarity matrix $S'_i$, the convolution process is as follows:
\begin{equation}
    F_{j,k}^{(l+1)}=\sum_{j=0}^{N_{max}} \sum_{k=0}^{N_{max}} ((S'_i)_j^{(l)}W_j^{(l)}+(S'_i)_k^{(l)}W_k^{(l)})
\end{equation}
where $i \in \{1,2\}$, $(S'_i)_j^{(l)},(S'_i)_k^{(l)}$ refers to the $j$-th row of the similarity matrix and the $k$-th column of the similarity matrix $S'_i$, respectively. 

Finally, after multiple convolution layers and pooling layers, a one-layer MLP is ultimately utilized to output the prediction value of GED, which is denoted as $\hat {S}$.

\subsection{Loss function}
The loss function is Mean Square Error (MSE), defined as:
\begin{equation}
    L=\frac{1}{\left| G_{i,j} \right| } \sum_{ \mathit{(i,j)} \in G_{i,j} }{(\hat {S}_{i,j}- y(G_i,G_j))}^2
\end{equation}
where $\left| G_{i,j} \right|$ refers to the number of graph pairs, $\hat {S}_{i,j}$ refers to the prediction of similarity score, and $y(G_i,G_j)$ refers to the ground-truth of GED value.

\section{Experiment}
We aim to answer the following questions in experiments.

\noindent \textbf{RQ1:} How does  the proposed  SEGMN  perform  compared 
to state-of-the-art graph similarity learning methods?

\noindent \textbf{RQ2:} Are the two proposed structure-enhanced modules in SEGMN effective?

\noindent \textbf{RQ3:} Can the structure perception matching module help to enhance other graph similarity learning methods?
 
\noindent \textbf{RQ4:} How can the structure perception matching module perceive structure difference?

\subsection{Experimental settings}
\noindent \textbf{Datasets.} Three benchmark real-world datasets, AIDS \cite{miss37, miss38,miss39}, LINUX \cite{miss45, miss46}, and IMDB \cite{miss40} are used to evaluate SEGMN. We follow GEDGNN \cite{miss25} to separate IMDB into two datasets composed of small graphs and large graphs respectively to evaluate the model on different scales of graphs. The details of the datasets are shown in Table 2.

\noindent \textbf{Baselines.} We compare SEGMN with two groups of baselines: (1) Simple GNN methods for graph-level matching, where GNNs like GCN \cite{miss10}, GIN \cite{miss34}, and GAT \cite{miss35} are for node embeddings, and mean pooling is conducted to get graph-level embedding for prediction; (2) State-of-the-art  methods with various node-level interaction, including SimGNN \cite{miss11}, GraphSim \cite{miss19}, EGSC-T \cite{miss26}, MGMN \cite{miss13}, ERIC \cite{miss24},  and NA-GSL \cite{miss15}.
More experiment details are available in Appendix C.
\subsection{Overall performance \textbf{(RQ1)}}
The performance of all methods is compared across the four datasets. For SEGMN,  GCN, GIN, and GraphSAGE are used as the backbone. The evaluation metrics detailed in Appendix C.1 include Mean Square Error (MSE), Spearman’s Rank Correlation Coefficient ($\rho$), Kendall’s Rank Correlation Coefficient ($\tau$), and precision at $k$ $(p@k)$, where MSE is the main metric and others show the matching extent of the ground-truth and the predicted results. 

As shown in Table 1, SEGMN achieves the best performance on MSE and most of the other metrics across all datasets. It is noteworthy that the nodes in LINUX and IMDB lack labels, causing the GED value to be entirely contingent on the structure similarity between the two graphs. Our approach demonstrates significant superiority over competitive methods on the two datasets, primarily due to its proficiency in perceiving the structure within inter-graph representation and cross-graph matching. Moreover, when applied to the IMDB dataset, which exhibits more edges than AIDS700 and LINUX, the dual embedding module equips each node with more adjacent edge information and achieves more improvement. SEGMN with different backbones is also competitive compared to other methods, indicating that more advanced GNN models can be adopted when handling different real-world application tasks.

\begin{table}
    \centering
    \renewcommand{\arraystretch}{0.95}
    \setlength{\tabcolsep}{3.5pt}
    \small
    \label{tab:booktable}
    \begin{tabular}{cccccc} 
        \toprule
        Datasets & Model & MSE($10^{-3}$) & $\rho$ & p@10 \\
        \midrule
             & node & 1.576 & 0.890 & 0.555 \\
             & edge & 1.670 & 0.898 & 0.539 \\
       AIDS700 & \textbf{DE}(node+edge) & 1.467 & 0.904 & 0.586 \\
       \cline{2-5}
             & node+\textbf{SPM} & 1.481 & 0.903 & 0.574 \\
             & edge+\textbf{SPM} & 1.631 & 0.905 & 0.584 \\
             & \textbf{DE}+\textbf{SPM}(\textbf{SEGMN}) & \textbf{1.386} & \textbf{0.910} & \textbf{0.597} \\
        \hline
             & node & 0.224 & 0.991 & 0.984 \\
             & edge & 0.136 & 0.992 & 0.987 \\
       LINUX & \textbf{DE}(node+edge) & 0.130 & 0.993 & 0.989 \\
       \cline{2-5}
             & node+\textbf{SPM} & 0.126 & 0.993 & 0.991 \\
             & edge+\textbf{SPM} & 0.118 & 0.993 & 0.991 \\
             & \textbf{DE}+\textbf{SPM}(\textbf{SEGMN}) & \textbf{0.094} & \textbf{0.994} & \textbf{0.994} \\
        \bottomrule
    \end{tabular}
    \caption{Ablation study. Firstly, we evaluate the performance of SEGMN using separate node embedding, edge embedding, and dual embedding (\textbf{DE}). Secondly, we study the difference when the structure perception matching module (\textbf{SPM}) is applied to each kind of embedding above.}
\end{table}

 \begin{figure*}
  \centering
  \includegraphics[width=0.72\textwidth]{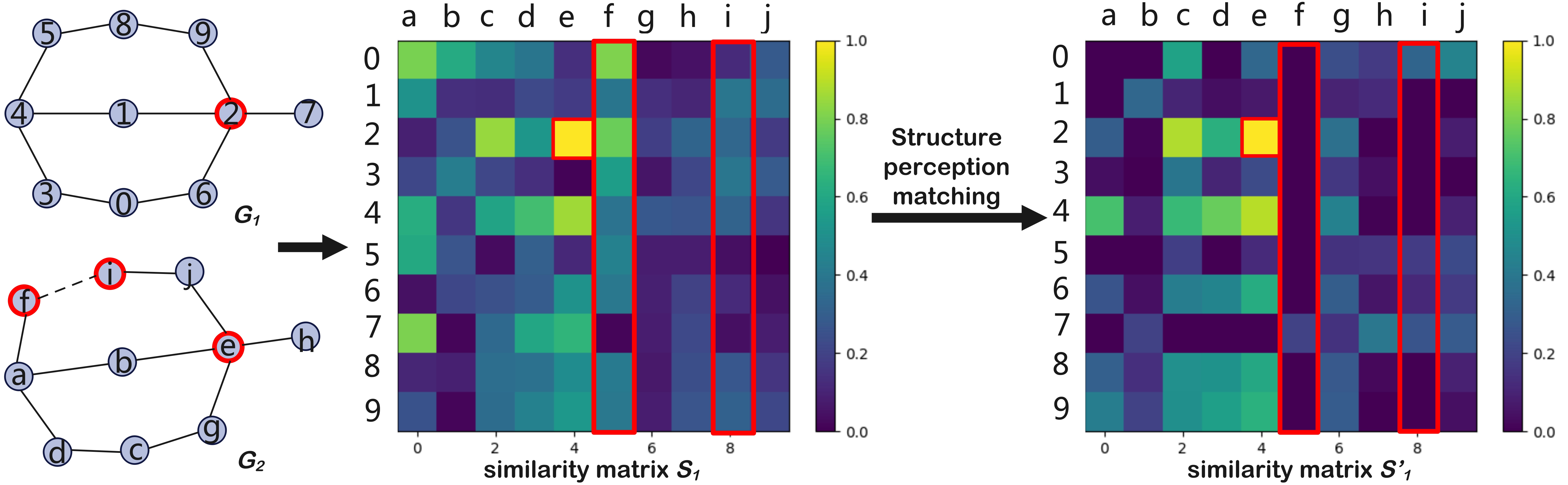}
  \caption{Case study.  The dotted line in $G_2$ shows the graph edit path, and if the edge is connected, they are the same. Similarity matrix $S_1$ is the similarity matrix from cross-graph interaction and $S'_1$ is the matrix after structure perception matching.}
  \label{fig:example} 
\end{figure*}

\subsection{Ablation study \textbf{(RQ2)}}
To evaluate the dual embedding module (\textbf{DE}) and structure perception matching module (\textbf{SPM}) in SEGMN, we conduct an ablation study on the labeled dataset AIDS700 and the unlabeled dataset LINUX.

Firstly, we study the effectiveness of the \textbf{DE} module. The node, edge, and DE (node+edge) in Table 3 represent the model using separate node embedding, separate edge embedding, and both respectively in the encoding process. We remove the SPM module to better observe the performance of different embedding strategies. The results show that the model using only node embedding performs better than that only using edge embedding on AIDS700, but the opposite is true on  LINUX. It is probably because the edge-enhanced GCN explores more structures that exactly determine the value of GED between unlabeled graphs. Overall, the dual embedding which benefits from both performs the best.

Then, the \textbf{SPM} module is added to the above models, and their performance is further improved. It indicates that conducting convolution in the assignment graph can help reconsider the similarity score from a more general perspective of the whole graph pair, and thus achieve structural matching. It's also noticeable that the SPM module brings more improvement to the model with separate node embedding, as the model with edge embedding has included intra-graph structure representation. Overall, SEGMN, which combines the DE and the SPM module, achieves the best performance on both labeled and unlabeled datasets.

\begin{table}
    \centering
    \renewcommand{\arraystretch}{0.95} 
    \setlength{\tabcolsep}{1pt}
    \small
    \label{tab:booktable}
    \begin{tabular*}{1\linewidth}{@{\extracolsep{\fill}}ccccccc}
        \toprule
         & \multicolumn{3}{c}{AIDS700} & \multicolumn{3}{c}{LINUX} \\
        
         Number & MSE  & $\rho$   & p@10  & MSE  & $\rho$   & p@10  \\
        \midrule
        0 & 2.014 & 0.839 & 0.401 & 0.762 & 0.953 & 0.956\\
        1 & 1.994 & 0.878 & 0.513 & 0.696 & 0.980 & 0.959\\
        2 & 1.958 & 0.879 & 0.515  & 0.664 & 0.981 & \textbf{0.963} \\
        3 & \textbf{1.875} & \textbf{0.883} & \textbf{0.523} & \textbf{0.565} & \textbf{0.982} & 0.961\\
        \midrule
         & \multicolumn{3}{c}{IMDB-small} & \multicolumn{3}{c}{IMDB-large} \\
        
         Number& MSE  & $\rho$   & p@10  & MSE  & $\rho$   & p@10  \\
        \midrule
		  0           &0.723  &0.940  & 0.976 &4.441  &0.615  & 0.558\\
		  1            &0.650  &0.959  &0.973 &4.340  &0.613  &0.564 \\
		  2      &0.594  &0.970  &0.982 &3.981  &0.685  &0.620  \\
		  3          &\textbf{0.536}  &\textbf{0.976}  &\textbf{0.984} &\textbf{3.469}  &\textbf{0.717}  &\textbf{0.632} \\
        \bottomrule
    \end{tabular*}
    \caption{Application of SPM module. The first column denotes the number of SPM modules applied to GraphSIM.}
\end{table}

\subsection{Portability and case study of the structure perception matching module \textbf{(RQ3\&RQ4)}}
\noindent \textbf{Portability.} The proposed SPM module can augment the similarity matrix with cross-graph structure information, so it can be plugged into any GSL method using similarity matrices \cite{miss11,miss19,miss30,miss33,miss15}.  We choose the classic method, GraphSim \cite{miss19}, for the portability experiment. In GraphSim, After each GCN layer,  a similarity matrix is attained based on the present node embeddings.  We add the SPM module after the first, the first and the second, and all three GCN layers respectively to observe the performance of GraphSim.

The results in Table 4 show diverse extents of improvement with different numbers of SPM modules. As the number of SPM modules increases, the performance of GraphSIM keeps improving. It's worth noting that adding the SPM module after the third layer of GCN brings more improvement than the other layers. It's reasonable as the third GCN provides a more comprehensive node representation to achieve better structural matching. The improvement on unlabeled datasets is relatively higher than that on the labeled dataset.

\noindent \textbf{Case study.} The case study demonstrates how the SPM module corrects the similarity score with structure information. The similarity matrix is normalized and presented in the form of a thermogram in Fig. 4. 
The thermogram illustrates that the entries in the similarity matrix $S_1$  display relatively modest differences. The differences are markedly accentuated after the SPM module, demonstrating its ability of distinguishing the structural difference between graphs. 

Moreover, in  $G_2$, $v^f$ and $v^i$ are not connected, so  $v^f, v^i, v^j, v^h$, are the branch nodes for the main structure ( the ring $v^a-v^b-v^e-v^g-v^c-v^d-v^a$), which means they are more similar with the branch node $v^7$ in $G_1$. This local structure is embodied in the assignment graph and then distinguished by the SPM module. Specifically, after SPM module, the similarity scores between $v^7$ and $v^f, v^i, v^j, v^h$ are all raised, while the similarity score between $v^f$ or $v^i$ and all nodes in $G_1$ are generally turned down (see red long boxes). The result is reasonable as $v^7$ and $v^f, v^i, v^j, v^h$ are all the branch nodes, their similarity scores should be raised,  while $v^f$ and $v^i$ are not connected, so their similarity with $G_1$ should be lowered. The similarity score between $v^2$ and $v^e$, which oughts to be the highest intuitively based on the graph structure, remains unchanged after the SPM module.

\section{Conclusion}

In this paper, we study the leverage of graph structures to learn graph similarity. Unlike existing models only using structures in message passing for node embedding, we focus on using structures to enhance both node embedding and cross-graph matching. Specifically,  a graph similarity learning framework based on a dual embedding learning module and structure matching perception module is proposed. On one hand, the node representation is enhanced by aggregating its adjacent edge features to generate dual embedding, providing each node with additional structure perspective. On the other hand, the structure perception matching module rectifies the similarity score of cross-graph node pairs by conducting convolution in the assignment graph. The abundant experiments verify the effectiveness and the advancement of our method compared with the SOTA methods.

\bibliography{aaai25}

\section{Reproducibility Checklist}

Unless specified otherwise, please answer “yes” to each question if the relevant information is described either in the paper itself or in a technical appendix with an explicit reference from the main paper. If you wish to explain an answer further, please do so in a section titled “Reproducibility Checklist” at the end of the technical appendix.

\paragraph{This paper:}

\begin{itemize}
    \item  Includes a conceptual outline and/or pseudocode description of AI methods introduced
    {\bf yes}
  \item Clearly delineates statements that are opinions, hypothesis, and speculation from objective facts and results
    {\bf yes}
   \item Provides well marked pedagogical references for less-familiare readers to gain background necessary to replicate the paper
    {\bf yes}
  
\end{itemize}

\paragraph{Does this paper make theoretical contributions?}
{\bf no}

\paragraph{Does this paper rely on one or more datasets?}
{\bf yes}

\begin{itemize}
    \item  A motivation is given for why the experiments are conducted on the selected datasets
    {\bf yes}
  \item 
All novel datasets introduced in this paper are included in a data appendix.
    {\bf yes}
   \item All novel datasets introduced in this paper will be made publicly available upon publication of the paper with a license that allows free usage for research purposes. 
    {\bf yes}
    \item All datasets drawn from the existing literature (potentially including authors’ own previously published work) are accompanied by appropriate citations. {\bf yes}
    \item All datasets drawn from the existing literature (potentially including authors’ own previously published work) are publicly available. {\bf yes}
    \item All datasets that are not publicly available are described in detail, with explanation why publicly available alternatives are not scientifically satisficing.  {\bf yes}
  
\end{itemize}

\paragraph{Does this paper include computational experiments?} {\bf yes}

\begin{itemize}
    \item  Any code required for pre-processing data is included in the appendix.
    {\bf yes}
  \item 
All source code required for conducting and analyzing the experiments is included in a code appendix.
    {\bf yes}
   \item All source code required for conducting and analyzing the experiments will be made publicly available upon publication of the paper with a license that allows free usage for research purposes.
    {\bf yes}
    \item All source code implementing new methods have comments detailing the implementation, with references to the paper where each step comes from {\bf yes}
    \item AIf an algorithm depends on randomness, then the method used for setting seeds is described in a way sufficient to allow replication of results.{\bf yes}
    \item This paper specifies the computing infrastructure used for running experiments (hardware and software), including GPU/CPU models; amount of memory; operating system; names and versions of relevant software libraries and frameworks.  {\bf yes}
    \item This paper formally describes evaluation metrics used and explains the motivation for choosing these metrics. {\bf yes}
    \item This paper states the number of algorithm runs used to compute each reported result. {\bf yes}
    \item Analysis of experiments goes beyond single-dimensional summaries of performance (e.g., average; median) to include measures of variation, confidence, or other distributional information. {\bf no}
    \item The significance of any improvement or decrease in performance is judged using appropriate statistical tests (e.g., Wilcoxon signed-rank). {\bf yes}
    \item This paper lists all final (hyper-)parameters used for each model/algorithm in the paper’s experiments. {\bf yes}
    \item This paper states the number and range of values tried per (hyper-) parameter during development of the paper, along with the criterion used for selecting the final parameter setting. {\bf yes}
  
\end{itemize}

\end{document}